\documentclass{article}

\PassOptionsToPackage{numbers}{natbib}

\usepackage[final]{nips_2018}

\usepackage[utf8]{inputenc} 
\usepackage[T1]{fontenc}    
\usepackage{hyperref}       
\usepackage{url}            
\usepackage{booktabs}       
\usepackage{amsfonts}       
\usepackage{nicefrac}       
\usepackage{microtype}      
\usepackage[pdftex]{graphicx}
\usepackage{xcolor}
\usepackage{multirow}
\usepackage{authblk}

\begin{document}  

\title{Deep Learning with Attention to Predict Gestational Age of the Fetal Brain}

\author[1]{Liyue Shen}
\author[2]{Katie Shpanskaya}
\author[1]{Edward Lee}
\author[2]{Emily McKenna}
\author[2]{Maryam Maleki}
\author[3]{Quin Lu}
\author[2]{Safwan Halabi}
\author[1]{John Pauly}
\author[2]{Kristen Yeom}
\affil[1]{Department of Electrical Engineering, Stanford University}
\affil[2]{Department of Radiology, Stanford University}
\affil[3]{Philips Healthcare North America}

\maketitle


\begin{abstract}

Fetal brain imaging is a cornerstone of prenatal screening and early diagnosis of congenital anomalies. Knowledge of fetal gestational age is the key to the accurate assessment of brain development. This study develops an attention-based deep learning model to predict gestational age of the fetal brain. The proposed model is an end-to-end framework that combines key insights from multi-view MRI including axial, coronal, and sagittal views. The model also uses age-activated weakly-supervised attention maps to enable rotation-invariant localization of the fetal brain among background noise. We evaluate our methods on the collected fetal brain MRI cohort with a large age distribution from 125 to 273 days. Our extensive experiments show age prediction performance with $R^2=0.94$ using multi-view MRI and attention.

\end{abstract}

\section{Introduction}
Fetal imaging has revolutionized prenatal care, allowing for early identification of structural and functional anomalies. Rigorous evaluation of fetal brain maturity and development is particularly critical due to the high prevalence of abnormalities (an estimated 3 in 1000 pregnancies)~\cite{phe2017}. Magnetic resonance imaging (MRI) has unrivaled diagnostic efficacy in detecting neurodevelopmental abnormalities. Despite its robust clinical utility, interpretation of fetal brain MRI remains a tremendous challenge for radiologists and clinicians due to the rapidly changing architecture of the normally maturing brain.

Knowledge of the gestational age of the fetus is crucial to the accurate interpretation of fetal brain MRI.  Despite its importance, our understanding of anatomical neurodevelopmental milestones in utero is limited to the gross evaluation of the cortical gyration (i.e. folding pattern of the brain’s cortical surface); and at this time, such coarse assessments are performed by a relatively “primitive” means of visual inspection and among a small pool of expert pediatric neuroradiologists at tertiary care fetal centers.

This study investigates the problem of predicting gestational age of fetal brain using multi-view MRI sequences. This study draws inspiration from the recent development of attention-based neural networks, a class of deep-learning methods that is capable of producing a spatial map that highlights regions of interest towards image object detection and interpretation~\cite{attention2018xray}. 
This study's contributions are as follows: (i) We demonstrate the proposed end-to-end framework is able to powerfully predict gestational age of fetal brains. (ii) We present experimental evidence that attention on the weakly-supervised activation maps can significantly improve the accuracy of these prediction from a baseline model without attention. (iii) The proposed multi-view prediction model largely improves the regression performance by incorporating useful information from different views.

\begin{figure*}[]
\centering
  \includegraphics[width=\linewidth]{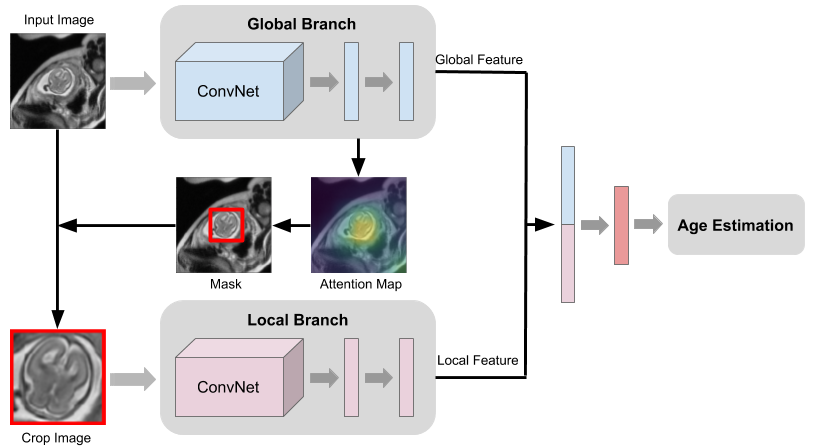}
  \caption{Model architecture for fetal brain age regression with attention mechanism and automatic mask inference.}
\label{fig:model}
\end{figure*}

\section{Methodology}

The model pipeline as shown in Figure ~\ref{fig:model} takes a 2D MRI scan image $X \in \mathbb{R}^{224\times224}$ and predicts the chronological age $y\in \mathbb{R}$. The 2D scan is selected from the center of a depth-wise stack of 2D MRI sequences for each patient. We define the objective as a regression problem to minimize $||y_{\mathrm{true age}} - y||_2^2$, where $y_{\mathrm{true age}}$ represents the true fetal brain ages and $y$ represents our model's predictions. For the model, we use residual networks~\cite{resnet} as the backbone model in the global branch for age regression. We use both ResNet-18 and ResNet-50 variant with different number of layers. We integrate three images (views) into our model: axial, sagittal, and coronal orientations. We train three models from each of these views, and combine these models together to form one final age prediction. 

Computational analysis for fetal brain images is extremely challenging because the random positions and orientations of fetal brain subregions are variable across patients. Furthermore, information that is unrelated to the fetus (mother's placenta and organs) act as clutter and may negatively affect performance. These reasons motivate the use of attention activation maps to crop out the part of $X$ that only pertains to the fetus.

\subsection{Attention to detect and localize the fetus} 
To extract image features pertaining to the fetal brain and to also filter out any clutter, we introduce an attention mechanism during the training of our model. As shown in Figure~\ref{fig:model}, an attention heatmap is extracted from the feature maps after the last stage of residual block with max intensity projection i.e. max pooling across channel dimension. 
The pixelwise values in the heatmap (attention map) highlight the local region where the network learns to pay higher attention. We use this attention map for automatic segmentation by thresholding pixel values higher than a fixed value, cropping out $X$ with a rectangular bounding box that captures the largest number of thresholded values with the minimum perimeter, and resizing (and interpolating) the cropped image $X_{\mathrm{cropped}}$ into the same size ($224\times224$) as the original image $X$. Finally, the cropped image $X_{\mathrm{cropped}}$ is fed into the local branch of our model (Figure~\ref{fig:model}). We study different ways of predicting the age: 1) using only the global branch (no attention and cropping), 2) using the local branch predictions from the masked inputs, 3) using the fusion branch by concatenating features from both local and global branches followed by fully-connected layer, and 4) averaging (instead of concatenating) global and local features.

\begin{figure*}[]
\centering
  \includegraphics[width=\linewidth]{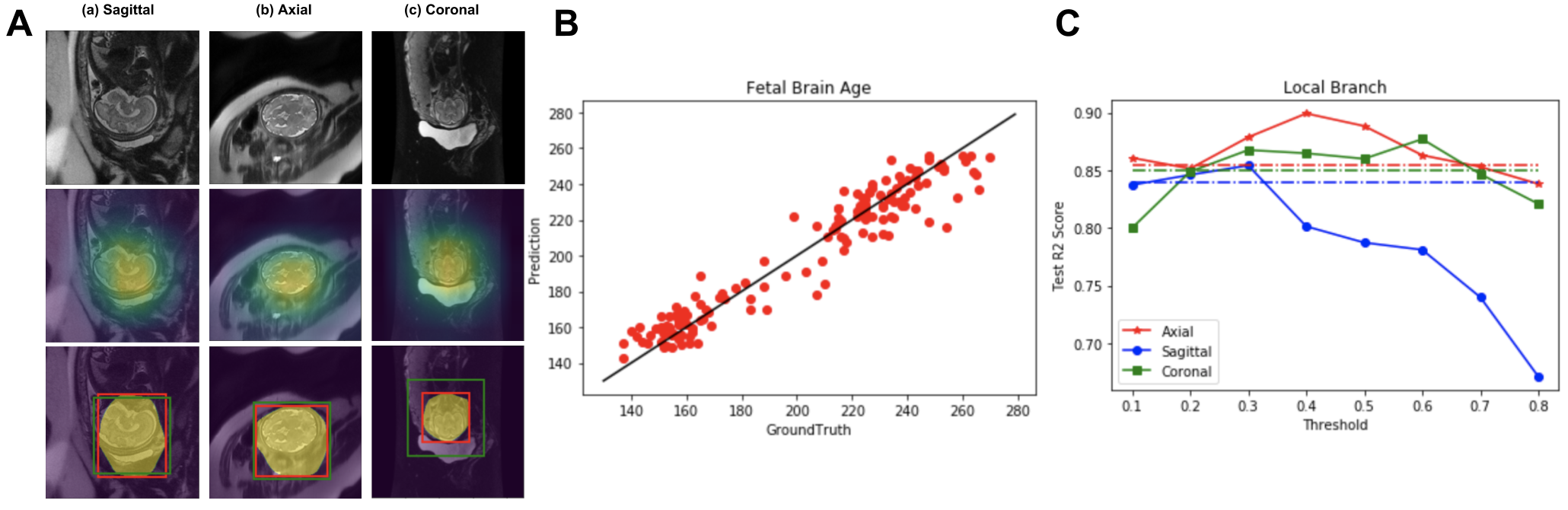}
  \caption{(A) Examples visualization of the attention mechanism. The top row shows examples of input images in three views: (a) Sagittal (b) Axial (c) Coronal. The middle row demonstrates the resulting attention maps extracted from feature maps in the last convolution layer. The bottom row shows the attention-guided automatic detention and sub-image cropping annotated as the red bounding boxes. (B) Visualization for regression performance on test set with attention-guided multi-view model. (C) Curve of test R2 scores v.s. threshold value in attention mechanism of local branch with ResNet-18 base model.
}
\label{fig:results}
\end{figure*}

Figure~\ref{fig:results} demonstrates learned visualization of attention heatmaps, mask inference, and automatic sub-image cropping from our pipeline. By visual inspection, we verify that our model learns the location of fetal brains in an unsupervised manner, given the small size of these brains with respect to the surrounding environment and with no groundtruth locations. This is strong evidence that the attention-aware models can learn local features from correct brain regions. Next, a sub-image is cropped from the whole input image (auto-segmentation). The thresholding value used to binarize and compute the bounding box is an important hyper-parameter. We discuss further in the next section. In experiments, without intensely working on tune this hyper-parameter, we set the threshold value as 0.3 for ResNet-18 and 0.4 for ResNet-50 respectively.

\section{Experiments}

\subsection{Dataset}

No public database of fetal MRI exists currently. We collected a robust database of 1927 fetal brain MRI from our clinical picture archiving and communication system (PACS). Each MRI was manually interpreted by an expert pediatric neuroradiologist to extract developmentally normal studies. Among those, the optimal T2-fast sequences in the three standard planes (axial, coronal, and sagittal) were identified. A total of 741 studies that had all three MRI planes were included in this study.
Fetal gestational age (in days) was calculated from the estimated date of delivery, in accordance with the current obstetrical guidelines and standard of care. These ages serve as ground truth labels and range from 125 to 273 days.
The entire dataset was split into training (70\%), validation(10\%), and test(70\%) sets.

\subsection{Results}
As the evaluation metric, the R2 score ranging from 0 to 1 and mean absolute error (MAE) is leveraged to show the quantized difference between model prediction and groundtruth labels.
For comparison, we show the results of different base architectures (ResNet-18 and ResNet-50) with various layers in Table~\ref{tab:r2} and Table~\ref{tab:mae}.
The results show that the increase of network depth benefits the final task of age regression, where more complicated feature representation are learned from deeper layers.
Besides, the ablative study also evaluates various comparison methods, such 
as single-view versus multi-view, with or without attention mechanism.
The quantization performance in the table indicates that both the attention mechanism and the multi-view learning are beneficial to the final prediction.
Specifically, in the single-view prediction, the axial and coronal planes can provide more useful information or more efficient image features to estimate fetal ages.
The combination of multi-view learning largely increases the regression accuracy compared to any single plane.
With the attention mechanism, the model can better understand the local image features such as brain shape or contour leading to a better evaluation performance, especially when the original regression performance without attention is not good enough.
Finally, by leveraging both the attention mechanism and multi-view learning, the state-of-the-art results are achieved with a R2 score of 0.94 and mean error of 6.8 days.

The qualitative visualization of this best result is demonstrated in Fig.~\ref{fig:results}(B), which demonstrates the qualitative visualization of regression performance on test dataset.
X-axis notes the groundtruth labels while Y-axis notes the model prediction ages for the corresponding samples.
The model-estimation results are expected to be aligned with groundtruth values as much as possible, which is annotated as the blue line of $y=x$.
Visual verification notes that the multi-view integration method with deep regression model can get a accurate performance for fetal brain age estimation.

\begin{table*}[]
\centering
\caption{\textbf{Evaluation: R2 Score}}
\label{tab:r2}
\begin{tabular}{l|l|lll|ll}
\toprule
\multicolumn{2}{c|}{\multirow{2}{*}{\textbf{Architecture}}}    & \multicolumn{3}{c|}{\textbf{Single-View}} & \multicolumn{2}{c}{\textbf{Multi-View}} \\ \cline{3-7}

\multicolumn{2}{c|}{}  & \textbf{Axial} & \textbf{Sagittal} & \textbf{Coronal} & \textbf{Average} & \textbf{Fusion} \\ \hline

\multirow{4}{*}{\textbf{ResNet-18}} 
& \textbf{Global Branch}& 0.8542 & 0.8391 & 0.8502  & 0.9107 & 0.9121 \\ \cline{2-7} 
& \textbf{Local Branch} & 0.8786 & 0.8541 & 0.8699  & \textbf{0.9283} & \color{red}\textbf{0.9348} \\ \cline{2-7}
& \textbf{Average Branch} & 0.8855 & \textbf{0.8620} & 0.8731  & 0.9250 & -  \\  \cline{2-7} 
& \textbf{Fusion Branch} & \textbf{0.8951} & 0.8566 & \textbf{0.8747}  & - &   0.9335  \\ 

\midrule
\multirow{4}{*}{\textbf{ResNet-50}} 
& \textbf{Global Branch}& 0.8878 & 0.8581 & 0.8644 & 0.9231 & 0.9269 \\ \cline{2-7} 
& \textbf{Local Branch} & 0.9141 & 0.8620 & 0.8814  & \textbf{0.9389} & \color{red} \textbf{0.9435} \\ \cline{2-7}
& \textbf{Average Branch} & \textbf{0.9175} & \textbf{0.8788} & \textbf{0.8975} & 0.9382  & -  \\  \cline{2-7} 
& \textbf{Fusion Branch}  & 0.9133  & 0.8768 & 0.8946  & - & 0.9420  \\ 
\toprule
\end{tabular}
\end{table*}

\begin{table*}[]
\centering
\caption{\textbf{Evaluation: Mean Absolute Error (MAE)}}
\label{tab:mae}
\begin{tabular}{l|l|lll|ll}
\toprule
\multicolumn{2}{c|}{\multirow{2}{*}{\textbf{Architecture}}}    & \multicolumn{3}{c|}{\textbf{Single-View}} & \multicolumn{2}{c}{\textbf{Multi-View}} \\ \cline{3-7}

\multicolumn{2}{c|}{}                                          & \textbf{Axial} & \textbf{Sagittal} & \textbf{Coronal} & \textbf{Average} & \textbf{Fusion} \\ \hline

\multirow{4}{*}{\textbf{ResNet-18}} 
& \textbf{Global Branch}& 11.0809 & 11.3324 & 11.1051  & 8.7452 & 8.6229 \\ \cline{2-7} 
& \textbf{Local Branch} & \textbf{9.9308} & 10.5633 & 10.2912  & \textbf{7.8268} & \color{red}\textbf{7.4677} \\ \cline{2-7}
& \textbf{Average Branch} & 10.0367 & \textbf{10.3415} & 9.9862  & 8.0102 & -  \\  \cline{2-7} 
& \textbf{Fusion Branch} & \textbf{9.9817} & 10.4094 & \textbf{9.8729}  & - &   7.6759  \\  
\midrule
\multirow{3}{*}{\textbf{ResNet-50}} 
& \textbf{Global Branch}& 9.4183 & 10.2276 & 10.4508 & 7.7450  & 7.4892 \\ \cline{2-7} 
& \textbf{Local Branch} & 8.6439 & 10.1733 & 9.1883  & \textbf{7.0172} & \textbf{6.9348} \\ \cline{2-7}
& \textbf{Average Branch} & \textbf{8.3740} & 9.4929 & \textbf{8.7842} & 7.0185  & -  \\  \cline{2-7} 
& \textbf{Fusion Branch}  & 8.6173  & 9.4712 & 9.1680  & - & \color{red} \textbf{6.8444}  \\ 
\toprule
\end{tabular}
\end{table*}

\subsection{Discussion}

To better understand how the threshold value in the attention mechanism influences the model learning, an analysis experiments with difference thresholds for ROI cropping was conducted.
The R2 score curve of test performance is shown in Fig.~\ref{fig:results}(C), which results from training the local branch ResNet-18 model using various thresholding values in attention mechanism.
From the curve, if the threshold value is too large or small, the learning prediction model will be damaged.
A smaller threshold will prefer to crop a larger sub-image and then push the local branch model to be closer to the global branch, which can not efficiently help extract local image features.
If the threshold is too large, the model will get worse, since almost no sub-images will be cropped for learning local regions and then the local branch model cannot learn anything.

\section{Conclusion}
This work proposes an end-to-end framework for efficient prediction of gestational age of fetal brain. 
Especially, experimental evidence shows that attention mechanism and multi-view learning can improve the accuracy of fetal brain age regression from a baseline model.
For the future work, 3D convolutional network will be investigated more.
External cohorts from outside institutions will also be used for further validation of this newly proposed method.

\end{document}